\documentclass[10pt,twocolumn,letterpaper]{article}
\usepackage{cvpr}              

\usepackage{graphicx}
\usepackage{amsmath}
\usepackage{amssymb}
\usepackage{booktabs}

\usepackage{array,multirow,graphicx}
\usepackage{float}
\usepackage{amsmath}
\usepackage{mathtools}
\usepackage{bbm}
\usepackage{hhline}
\usepackage{boldline}
\usepackage{tabularx}
\usepackage{comment}
\usepackage{color}
\usepackage{amssymb}
\usepackage{booktabs}
\usepackage{hhline}
\usepackage{multicol}
\usepackage{microtype}      
\usepackage{xcolor}         
\usepackage{hhline}
\usepackage{bbm}
\usepackage{boldline}
\usepackage{amssymb}
\usepackage{algorithm}
\usepackage{algpseudocode}
\usepackage{pifont}
\usepackage{soul} 
\newcommand{\cmark}{\ding{51}}%
\usepackage{subcaption}

\usepackage[accsupp]{axessibility}  

\usepackage[pagebackref,breaklinks,colorlinks]{hyperref}

\usepackage[capitalize]{cleveref}

\crefname{section}{Sec.}{Secs.}
\Crefname{section}{Section}{Sections}
\Crefname{table}{Table}{Tables}
\crefname{table}{Tab.}{Tabs.}

\def\cvprPaperID{6761} 
\def\confName{CVPR}
\def\confYear{2023}

\newcommand{\Skip}[1]{}

\begin{document}

\title{Query-Dependent Video Representation \\ for Moment Retrieval and Highlight Detection}

\newcommand\blfootnote[1]{%
  \begingroup
  \renewcommand\thefootnote{}\footnote{#1}%
  \addtocounter{footnote}{-1}%
  \endgroup
}
\author{
WonJun Moon$^{1,\ast}$, Sangeek Hyun$^{1,\ast}$, SangUk Park$^{2}$, Dongchan Park$^{2}$, Jae-Pil Heo$^{1,\star}$\\
$^{1}$Sungkyunkwan University, $^{2}$Pyler\\
{\tt\small \{wjun0830, hsi1032, jaepilheo\}@g.skku.edu,} {\tt\small \{psycoder, cto\}@pyler.tech}
}

\maketitle

\begin{abstract}
Recently, video moment retrieval and highlight detection~(MR/HD) are being spotlighted as the demand for video understanding is drastically increased.
The key objective of MR/HD is to localize the moment and estimate clip-wise accordance level, i.e., saliency score, to the given text query.
Although the recent transformer-based models brought some advances, we found that these methods do not fully exploit the information of a given query.
For example, the relevance between text query and video contents is sometimes neglected when predicting the moment and its saliency.
To tackle this issue, we introduce Query-Dependent DETR~(QD-DETR), a detection transformer tailored for MR/HD.
As we observe the insignificant role of a given query in transformer architectures, our encoding module starts with cross-attention layers to explicitly inject the context of text query into video representation.
Then, to enhance the model's capability of exploiting the query information, we manipulate the video-query pairs to produce irrelevant pairs. 
Such negative~(irrelevant) video-query pairs are trained to yield low saliency scores, which in turn, encourages the model to estimate precise accordance between query-video pairs.
Lastly, we present an input-adaptive saliency predictor which adaptively defines the criterion of saliency scores for the given video-query pairs.
Our extensive studies verify the importance of building the query-dependent representation for MR/HD.
Specifically, QD-DETR outperforms state-of-the-art methods on QVHighlights, TVSum, and Charades-STA datasets. 
Codes are available at \href{github.com/wjun0830/QD-DETR}{github.com/wjun0830/QD-DETR}.
\blfootnote{
$^*$ Equal contribution} 
\blfootnote{
$^\star$ Corresponding author
}
\end{abstract}

\begin{figure}
    \centering
    \includegraphics[width=1\linewidth]{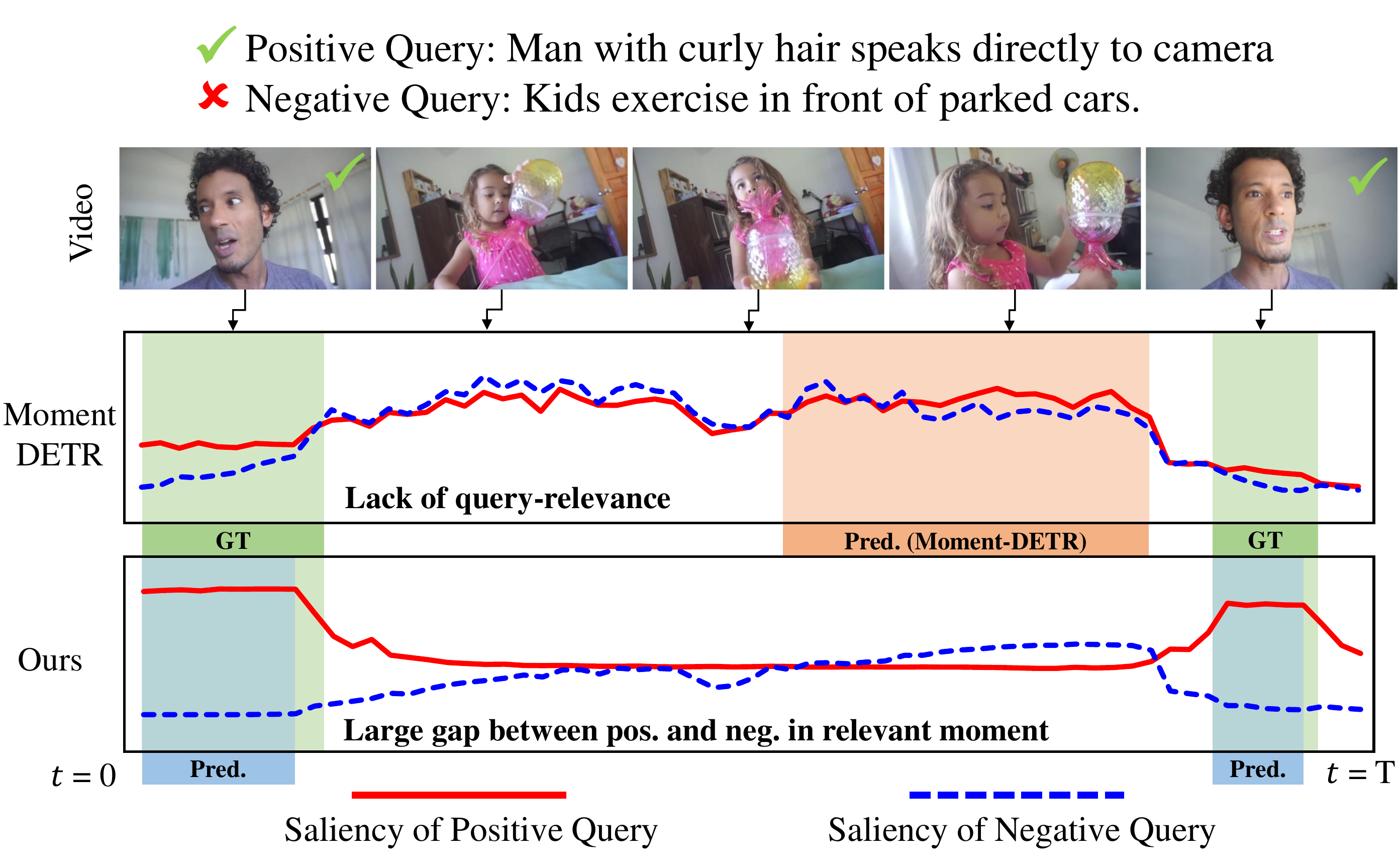}
    \vspace{-0.6cm}
    \caption{
        Comparison of highlight-ness~(saliency score) when relevant and non-relevant queries are given.
        We found that the existing work only uses queries to play an insignificant role, thereby may not be capable of detecting negative queries and video-query relevance; saliency scores for clips in ground-truth~(GT) moments are low and equivalent for positive and negative queries.
        On the other hand, query-dependent representations of QD-DETR result in corresponding saliency scores to the video-query relevance and precisely localized moments.
}
    \label{fig:motivation_ex}
\end{figure}

\section{Introduction}
Along with the advance of digital devices and platforms, video is now one of the most desired data types for consumers~\cite{apostolidis2021video,wu2017deep}.
Although the large information capacity of videos might be beneficial in many aspects, e.g., informative and entertaining, inspecting the videos is time-consuming, so that it is hard to capture the desired moments~\cite{anne2017localizing,apostolidis2021video}. 

Indeed, the need to retrieve user-requested or highlight moments within videos is greatly raised.
Numerous research efforts were put into the search for the requested moments in the video~\cite{anne2017localizing, gao2017tall, liu2015multi, escorcia2019temporal} and summarizing the video highlights~\cite{zhang2016video, mahasseni2017unsupervised, badamdorj2022contrastive, wei2022learning}.
Recently, Moment-DETR~\cite{momentdetr} further spotlighted the topic by proposing a QVHighlights dataset that enables the model to perform both tasks, retrieving the moments with their highlight-ness, simultaneously.

When describing the moment, one of the most favored types of query is the natural language sentence~(text)\cite{anne2017localizing}. 
While early methods utilized convolution networks~\cite{zhang2020learning, gao2021fast, wang2020temporally}, recent approaches have shown that deploying the attention mechanism of transformer architecture is more effective to fuse the text query into the video representation.
For example, Moment-DETR~\cite{momentdetr} introduced the transformer architecture which processes both text and video tokens as input by modifying the detection transformer~(DETR), and UMT~\cite{umt} proposed transformer architectures to take multi-modal sources, e.g., video and audio. 
Also, they utilized the text queries in the transformer decoder.
Although they brought breakthroughs in the field of MR/HD with seminal architectures, they overlooked the role of the text query.
To validate our claim, we investigate the Moment-DETR~\cite{momentdetr} in terms of the impact of text query in MR/HD~(Fig.\ref{fig:motivation_ex}).
Given the video clips with a relevant positive query and an irrelevant negative query, we observe that the baseline often neglects the given text query when estimating the query-relevance scores, i.e., saliency scores, for each video clip.

To this end, we propose Query-Dependent DETR~(QD-DETR) that produces query-dependent video representation.
Our key focus is to ensure that the model's prediction for each clip is highly dependent on the query.
First, to fully utilize the contextual information in the query, we revise the transformer encoder to be equipped with cross-attention layers at the very first layers.
By inserting a video as the query and a text as the key and value of the cross-attention layers, our encoder enforces the engagement of the text query in extracting video representation.
Then, in order to not only inject a lot of textual information into the video feature but also make it fully exploited, we leverage the negative video-query pairs generated by mixing the original pairs.
Specifically, the model is learned to suppress the saliency scores of such  negative~(irrelevant) pairs.
Our expectation is the increased contribution of the text query in prediction since the videos will be sometimes required to yield high saliency scores and sometimes low ones depending on whether the text query is relevant or not.
%
Lastly, to apply the dynamic criterion to mark highlights for each instance, we deploy a saliency token to represent the entire video and utilize it as an input-adaptive saliency criterion. 
With all components combined, our QD-DETR produces query-dependent video representation by integrating source and query modalities.
This further allows the use of positional queries~\cite{dabdetr} in the transformer decoder.
Overall, our superior performances over the existing approaches validate the significance of the role of text query for MR/HD.
\section{Related Work}
\subsection{Moment Retrieval and Highlight Detection}
MR is the task of localizing the moment relevant to the given text description.
Popular approaches are modeling the cross-modal interaction between text query-video pair~\cite{zhang2020span,yuan2019semantic,lu2019debug} or understanding the context of the temporal relation among video clips~\cite{anne2017localizing,zhang2020learning}.
On the other hand, 
TVT~\cite{lei2020tvr} exploited the additional data, i.e., subtitle, to capture the moment, and FVMR~\cite{gao2021fast} enhanced the model in terms of inference speed for efficient MR.

Different from the MR, HD aims to measure the clip-wise importance level of the given video~\cite{sun2014ranking, yao2016highlight}.
Due to its popularity and applicability, HD can be divided into several branches.
From the perspective of annotation, we can categorize HD into supervised, weakly supervised, and unsupervised HD.
Supervised HD~\cite{gygli2016video2gif, xu2021cross, sun2014ranking} utilizes fine-grained highlight scores, which are very expensive to collect and annotate~\cite{xiong2019less}.
On the other hand, weakly supervised HD~\cite{cai2018weakly, panda2017weakly, xiong2019less} learns to detect segments as highlights 
with video event labels, and finally, unsupervised HD~\cite{badamdorj2022contrastive,mahasseni2017unsupervised, khosla2013large, rochan2018video} does not require any annotations.
Also, while the task is often implemented only with the video, there are works to employ the extra data modalities.
Generally, multi-modality was taken into account by using the natural language query to find the desired thumbnail~\cite{liu2015multi} and using additional sources, i.e., audio, to predict highlights~\cite{ye2021temporal,badamdorj2021joint}.
\begin{figure*}
    \centering
    \includegraphics[width=\linewidth]{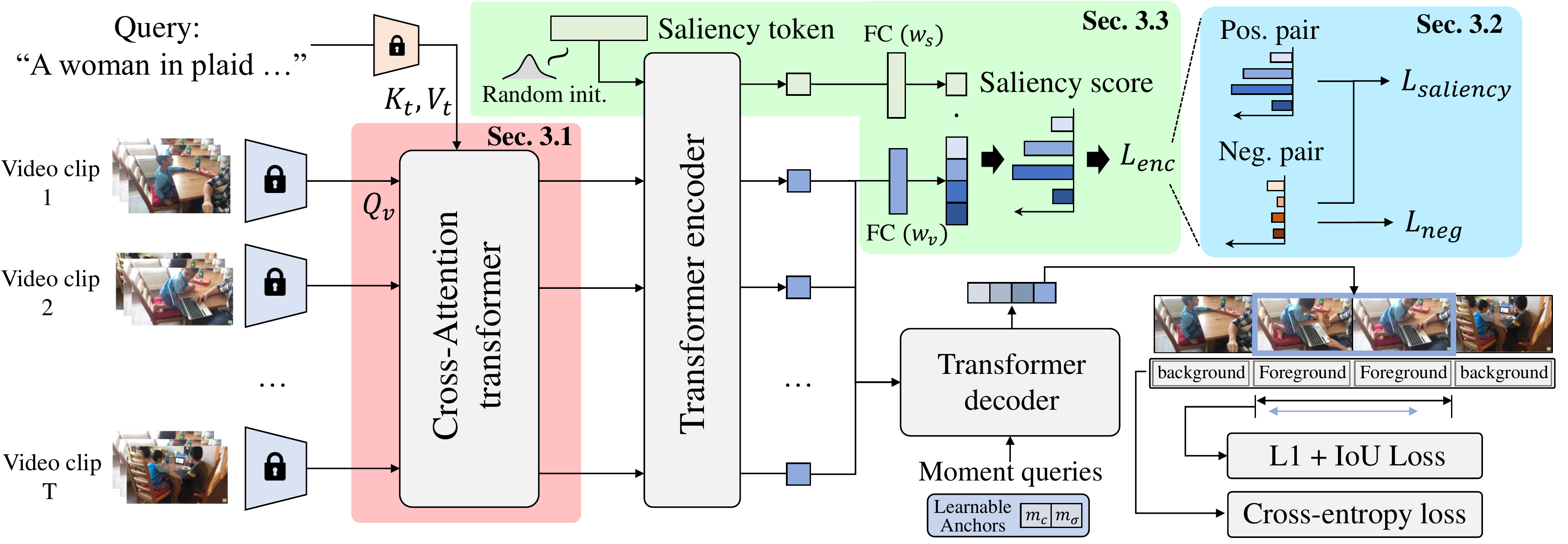}
    \caption{
    Overview of the proposed QD-DETR architecture.
    Given a video and text query, we first extract video and text features from the frozen backbones.
    These video and text features are forwarded into the cross-attention transformer~(Sec.~\ref{Sec.CrossAttentiveTransformerEncoder}). This process ensures the consistent contribution of text queries to the video tokens and together with negative pair learning~(Sec.~\ref{Sec.VideoDiscriminativeQueryInformation}), builds query-dependent video representation.
    Then, accompanied by the saliency token~(Sec.~\ref{Sec.SaliencyToken}), video tokens are given to the transformer encoder.
    In this procedure, the saliency token is transformed into the adaptive saliency prediction criteria.
    The outputs of the encoder are then, processed to compute losses for both HD and MR.
    Specifically, the encoder's output tokens are directly projected to saliency scores and optimized for HD, and also provided to the transformer decoder with the learnable moment queries to estimate the query-described moments.
    Finally, losses for MR are computed by the discrepancy between predicted and their corresponding GT moments.
    }
    \label{fig:overall architecture}
\end{figure*}


Although the MR/HD share a common objective to localize or discover the desired part of the given video, they have been studied separately.
To handle these tasks at once, Moment-DETR~\cite{momentdetr} proposed the QVHighlights dataset, which contains a human-written text query and its corresponding moment with clip-level saliency labels.
They also introduced the modified version of detection transformer~(DETR~\cite{detr}) to localize the query-relevant moments and their saliency scores.
Following them, UMT~\cite{umt} focused on processing multi-modal data by utilizing both video and audio features.
Different from recent works deploying transformer architectures, here we concentrate on producing a query-dependent representation with the transformer.

\subsection{Detection Transformers}
DETR~\cite{detr}, an end-to-end object detector based on vision transformers, is one of the very recent works that 
utilize the transformer architectures for computer vision~\cite{dosovitskiy2020image, touvron2021training}.
Although DETR suffered from slow convergence, it simplifies the prediction process by eliminating the need for anchor generation and non-maximum suppression.
Since then, along with the advance in DETR~\cite{dai2021dynamic, zhu2020deformable, li2022dn}, DETR-like architectures have been popular in downstream tasks in both the image~\cite{cheng2022masked, cheng2021per, huang2022monodtr, zhang2022monodetr} and video domains~\cite{yang2022tubedetr, momentdetr}.
Some of these works focused on analyzing the role of the decoder query and discovered that using the positional information speeds up the training and also enhances the detection performance~\cite{conditionaldetr, dabdetr}.
On the other hand, there are trials to extend the application of DETR on multi-modal data~\cite{kamath2021mdetr, momentdetr}, especially dealing with the query from different modalities, i.e., text, for detection~(or retrieval).
They generally handle the multi-modal data 
by simply forwarding them together to the transformer.
In this paper, we also focus on handling the multi-modal data based on DETR-like architecture.
However, different from the aforementioned techniques, we concentrate on the query-dependency of the prediction results.


\section{Query-Dependent DETR}
Moment retrieval and highlight detection have the common objective to find preferred moments with the text query.
Given a video of $L$ clips and a text query with $N$ words, we denote their representations as $\{v_1, v_2, ..., v_L\}$ and $\{ t_1, t_2, ..., t_{N} \}$ extracted by frozen video and text encoders, respectively.
With these representations, the main objective is to localize the center coordinate $m_c$ and width $m_\sigma$ within the video and rank the highlight score~(saliency score) $\{ s_1, s_2, ..., s_{L} \}$ for each clip.
A straightforward approach to utilize transfomer~\cite{vaswani2017attention, dosovitskiy2020image} for the MR is to make a moment-wise prediction as a set of clips~\cite{momentdetr}, or generating the moment according to the clip-wise predictions~\cite{umt}.
To exploit the multi-modal information, e.g., video and text query, they either simply concatenated the features across the modalities or inserted the texts to form the moment query to the transformer decoder. 
However, we claim that the relationship between the video and text query should be carefully considered rather than a simple concatenation since MR/HD requires every video clip to be conditionally assessed with the text queries.

Our overall architecture is described in Fig.~\ref{fig:overall architecture}, following the design of concrete baseline, Moment-DETR~\cite{momentdetr}.
Given a video and query representation extracted from fixed backbones, QD-DETR first transforms the video representation to be query-dependent using cross-attention layers. To further enhance the query-awareness of video representations, we incorporate irrelevant video-query pairs with a low saliency for the learning objective. Then, along with the transformer encoder-decoder architectures, the saliency token is defined that turns into an adaptive saliency predictor when attended by the specific video instance.



\subsection{Cross-Attentive Transformer Encoder}
\label{Sec.CrossAttentiveTransformerEncoder}
In this subsection, we use italic letters to represent {\it query}, {\it key}, and {\it value} of the cross-attention layers.
The key objective of the encoder for MR/HD is to produce clip-wise representations equipped with information regarding the degree of query-relevance since these features are directly used for retrieving the query-matched moments and predicting clip-wise saliency scores.
However, the encoding process of existing works may not ensure the query conditioning on every clip.
For example, Moment-DETR~\cite{momentdetr} naively concatenated the video with the query 
for input to the self-attention layers, which may result in an insignificant role of the query if the high similarities among the video clips overwhelm the contribution of the text query.
On the other hand, UMT~\cite{umt} utilizes the text query only for the synthesis of a moment query in the transformer decoder so thus resulting video representations are not associated with the text query.

To take the textual contexts into every video clip representation, we deploy cross-attention layers between the source and the query modalities at the very first layers of the encoder.
This ensures the consistent contribution of the query, thereby extracting query-dependent video representation.
In detail, whereas the {\it query} for cross-attention layers is prepared by projecting the video clips as $Q_v = [~p_q(v_1), ..., p_q(v_L)~]$, the {\it key} and {\it value} are computed with the query text features as $K_t = [~p_k(t_1), ..., p_k(t_N)~]$ and $V_t = [~p_v(t_1), ..., p_v(t_N)~]$.
$p_q(\cdot)$, $p_k(\cdot)$, and $p_v(\cdot)$ are projection layers for {\it query}, {\it key}, and {\it value}.
Then, the cross-attention layer operates as follows:
\begin{eqnarray}
    \label{eqn 1 cross attention}
    \text{Attention}(Q_v, K_t, V_t) = \text{softmax}(\frac{Q_vK_t^T}{\sqrt{d}})V_t,
\end{eqnarray}
where $d$ is the dimension of the projected {\it key}, {\it value}, and {\it query}.
Since the softmax scores are distributed only over the query elements, video clips are expressed with the weighted sum of the text queries in proportion to the similarity to texts.
Attention scores are then projected through MLP and integrated into the original video representations as the typical transformer layers. 
For the rest of the paper, we define the query-dependent video tokens, i.e., the output of cross-attention layers, as $X = \{ x_v^1, x_v^2, ..., x_v^L \}$.


\begin{figure}[t]
      \centering
        \includegraphics[width=1.05\linewidth]{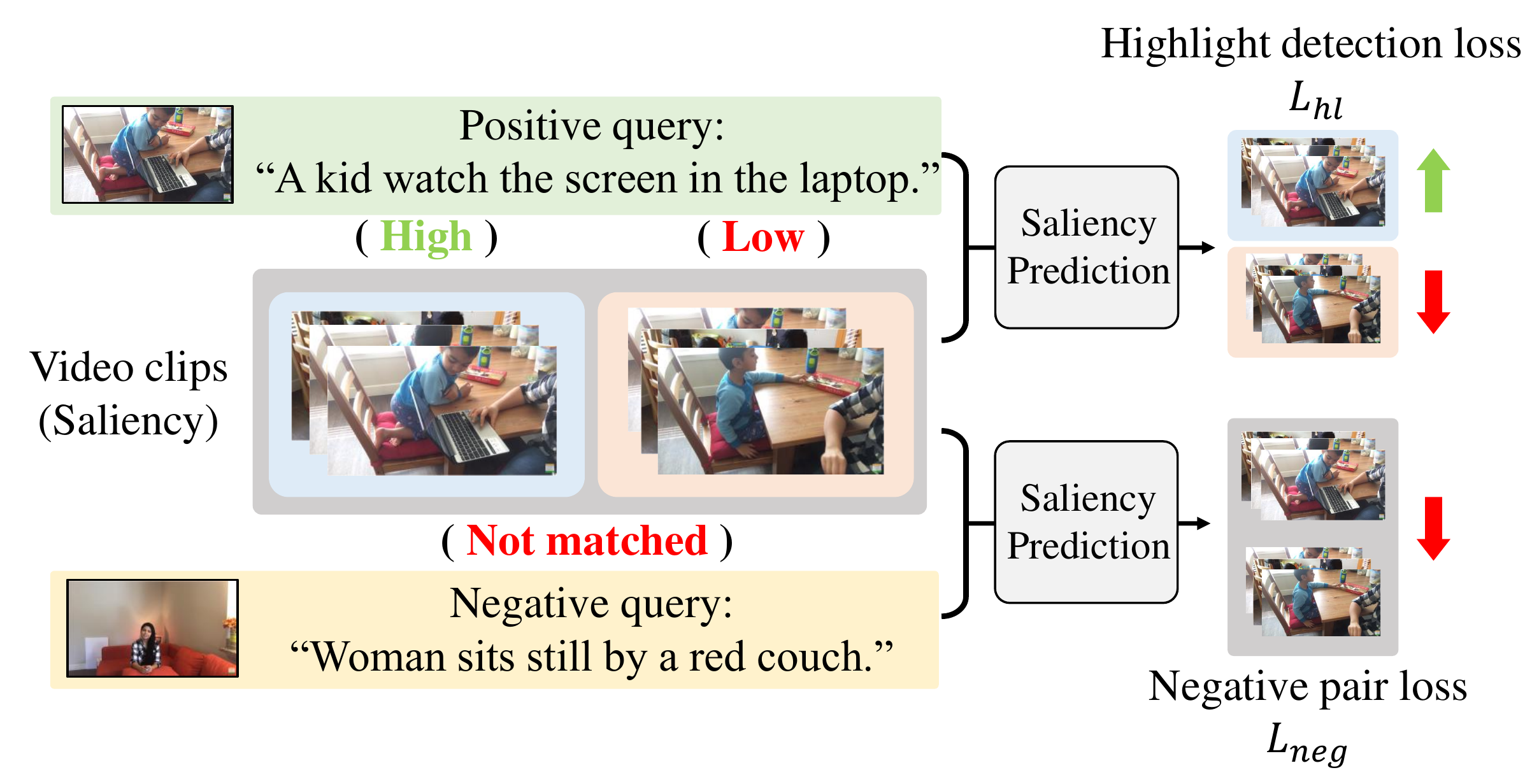}
        \vspace{-0.8cm}
        \caption{
            Illustration of negative pair learning. Typical HL loss is defined only with a positive video-query pair, which is insufficient to learn various degrees of query-relevance. 
            On the other hand, our negative pair learning enforces the model to yield different scores for a video depending on the query and to be learned to suppress saliency scores for the negative query.
        }
    \label{fig:neg_pair}
\end{figure}

\subsection{Learning from Negative Relationship}
\label{Sec.VideoDiscriminativeQueryInformation}
While the cross-attention layers explicitly fuse the video and query features for intermediate video clip representations to engage the query information in an architectural way, we argue that given video-text pairs lack diversity to learn the general relationship.
For instance, many consecutive clips in a single video often share similar appearances, and the similarity to a specific query will not be highly distinguishable, thereby, the text query may not much affect the prediction.

Thus, we consider the relationships between irrelevant pairs of videos 
inspired by many recognition practices~\cite{he2016deep, lin2017focal, li2020dividemix, moon2022difficulty} that learn discriminative features across different categories.
To implement such relationships, we define given training video-query pairs as positive pairs and mix the video and query from different pairs to construct negative pairs.
Fig.~\ref{fig:neg_pair} illustrates the ways to augment such negative pairs and utilize them with positive pairs in training.
While the video clips in positive pairs are trained to yield segmented saliency scores according to the query-relevance, irrelevant negative video-query pairs are enforced to have the lowest saliency scores.
Formally, the loss function for suppressing the saliency of negative pairs $x_v^{\text{neg}}$ are expressed as follows:
\begin{eqnarray}
    \label{eq.negative learning}
    L_{\text{neg}} = - \log(1 - S(x_v^{\text{neg}})),
\end{eqnarray}
where $S(\cdot)$ is the saliency score predictor.
This training scheme can also prevent the model from predicting the moments and highlights solely based on the inter-relationship among video clips without consideration of the query-relevance since the same video instance should be predicted differently depending on whether the positive or negative query is given.

\subsection{Input-Adaptive Saliency Predictor}
\label{Sec.SaliencyToken}
Naive implementation for saliency predictor $S(\cdot)$ would be stacking one or more fully-connected layers. 
However, such a general head provides identical criteria for the saliency prediction of every video-query pair, neglecting the diverse nature of video and natural language query pairs. This violates our key idea to extract query-dependent video representation.


Thus, we define the saliency token $x_s$ to be utilized as an input-adaptive saliency predictor.
Briefly, the saliency token is a randomly initialized learnable vector that becomes an input-adaptive predictor when added to the sequence of encoded video tokens and projected through the transformer encoder. 
To illustrate, as shown in Fig.~\ref{fig:overall architecture}, we first concatenate the saliency token with the query-dependent video tokens $X$.
We process these tokens to the transformer encoder which makes the saliency token to be re-organized with the input-dependent contexts.
Consequently, saliency and video tokens are projected by a corresponding single fully-connected layer with weights, $w_{s}$ and $w_{v}$, respectively, where their scaled-dot product becomes the saliency scores.
Formally, saliency score $S(x_v^i)$ is computed as follows:
\begin{eqnarray}
    \label{eq. saliency score prediction}
     S(x_v^i) = \frac{w_s^T x_s \cdot w_v^T x_v^i}{\sqrt{d}},
\end{eqnarray}
where $d$ is the channel dimension of projected tokens.



\subsection{Decoder and Objectives}
\paragraph{Transformer Decoder.}
Recently, understanding the role of the query in the detection transformer is being spotlighted~\cite{conditionaldetr, dabdetr}. 
It is verified that designing the query with the positional information helps not only for acceleration of training but also for enhancing accuracy.
Yet, it is hard to directly employ these studies in tasks handling multi-modal data, e.g., MR/HD, since multi-modal data often have different definitions of position; the position can be understood as time in the video and word order in the text.

On the contrary, our architectural design eliminates the need to feed the text query to the decoder since the query information is already taken into the video representations.
To this end, we modify the 2D dynamic anchor boxes~\cite{dabdetr} to represent 1D moments in the video.
Specifically, we utilize the center coordinate $m_c$ and the duration $m_\sigma$ of the moments to design the queries.
Similarly to the previous way in the image domain, we pool the features around the center coordinate and modulate the cross-attention map with the moment duration. Then, the coordinates and durations are layer-wisely revised.

\paragraph{Loss Functions.}
Training objectives for QD-DETR include loss functions for MR/HD, respectively.
First, objective functions for MR, in which the key focus is to locate the desired moments, are adopted from the baseline~\cite{momentdetr}.
Moment retrieval loss $L_{\text{mr}}$ measures the discrepancy between the GT moment and the predicted counterpart.
It consists of a $L1$ loss and a generalized IoU loss $L_{\text{gIoU}}(\cdot)$ from previous work~\cite{rezatofighi2019generalized} with minor modification to localize temporal moments.
Additionally, the cross-entropy loss is used to classify the predicted moments as  $\hat{y}$ either to foreground and background by $L_{\text{CE}} = - \sum_{y \in Y}^{} y \log ( \hat{y} )$ where $\{\text{fg}, \text{bg}\} \subset Y$.
Thus, $L_{\text{mr}}$ is defined as follows:
\begin{eqnarray}
    \label{eqn moment localization loss}
    &L_{\text{mr}} = \lambda_{L1} ||m - \hat{m}|| + \lambda_{\text{gIoU}} L_{\text{gIoU}}(m, \hat{m}) + \lambda_{\text{CE}} L_{\text{CE}},
\end{eqnarray}
where $m$ and $\hat{m}$ are ground-truth moment and its correspond prediction containing center coordinate $m_c$ and duration $m_\sigma$. 
Also, $\lambda_{*}$ are hyperparameters for balancing the losses.

Loss functions for HD are to estimate the saliency score.
It comprises two components; margin ranking loss $L_{\text{margin}}$ and rank-aware contrastive loss $L_{\text{cont}}$. 
Following \cite{momentdetr}, the margin rank loss operates with two pairs of high-rank and low-rank clips.
To be specific, the high-rank clips are ensured to retain higher saliency scores than both the low-rank clips within the GT moment and the negative clips outside the GT moment.
In short, $L_{\text{margin}}$ is defined as: 
\begin{eqnarray}
    \label{eqn margin ranking loss}
    L_{\text{margin}} = \text{max}(0, \Delta + S(x_{}^{\text{low}}) - S(x_{}^{\text{high}})) 
\end{eqnarray}
where $\Delta$ is the margin, $S(\cdot)$ is the saliency score estimator, and $x^{\text{high}}$ and $x_{}^{\text{low}}$ are video tokens from two pairs of high and low-rank clips, respectively.
In addition to margin loss which only indirectly guides the saliency predictor, we employ rank-aware contrastive loss~\cite{hoffmann2022ranking} to learn the precisely segmented saliency levels with the contrastive loss.
Given the maximum rank value $R$, each clip in the mini-batch has a saliency score lower than $R$.
Then, we iterate the batch for $R$ times, each time utilizing the samples with higher saliency scores than the iteration index~($r \in \{0, 1, ..., R-1\}$) to build the positive set $X^{\text{pos}}_r$.
Samples with a lower rank than the iteration index are included in the negative set $X^{\text{neg}}_r$.
Then, the rank-aware contrastive loss $L_{\text{cont}}$ is defined as: 
\begin{eqnarray} 
    \label{eqn contrastive loss}
    L_{\text{cont}}=-\sum_{r=1}^{R}\text{log}\frac{
    \sum_{x\in X_r^\text{pos}}\text{exp}(S(x)/\tau)}
    {
    \sum_{x\in (X_r^\text{pos} \cup X_r^\text{neg})}\text{exp}(S(x)/\tau)
    }
\end{eqnarray}
where $\tau$ is a temperature scaling parameter.
Note that, $X^{\text{neg}}_r$ also include all clips in negative pairs $x_v^{\text{neg}}$ defined in Sec.~\ref{Sec.VideoDiscriminativeQueryInformation}.
Finally with margin loss and rank-aware contrastive loss, $L_{\text{hl}}$ and total loss function $L_{\text{total}}$ are defined as follows:
\begin{eqnarray}
    \label{eqn moment localization loss}
    &L_{\text{hl}} = \lambda_{\text{margin}} L_{\text{margin}} + \lambda_{\text{cont}} L_{\text{cont}}, \\
    &L_{\text{total}} = L_{\text{hl}} 
    + L_{\text{mr}} + \lambda_{\text{neg}} L_{\text{neg}}.
\end{eqnarray}








\section{Evaluation}
\label{Sec.Experiments}

\begingroup
\setlength{\tabcolsep}{7pt} 
\renewcommand{\arraystretch}{1} 
\begin{table*}[]
	\centering
	{\small 
	\vspace{-0.1cm}  %
	\caption{Performance comparison on QVHighlights \textit{test} split. V and A in the Src column denote video and audio, respectively, representing the modalities of the source data. 
 Our experiments are averaged over five runs and `$\pm$' denotes the standard deviation.}
	\label{table_QVHighlight}
        \begin{tabular}{l|c|ccccccc}
        \hlineB{2.5}
        \multicolumn{1}{c}{\multirow{3}{*}{Method}} & \multirow{3}{*}{Src} & \multicolumn{5}{c}{MR}                                                             & \multicolumn{2}{c}{HD}                        \\ \cline{3-9} 
        \multicolumn{1}{c}{} & \multicolumn{1}{c|}{} & \multicolumn{2}{c}{R1}          & \multicolumn{3}{c}{mAP}                          & \multicolumn{2}{c}{\textgreater{}= Very Good} \\ \cline{3-9} 
        \multicolumn{1}{c}{} & \multicolumn{1}{c|}{} & @0.5           & @0.7           & @0.5           & @0.75          & Avg.           & mAP                   & HIT@1                 \\ \hlineB{2.5}
        BeautyThumb~\cite{song2016click}               & V                  & -              & -              & -              & -              & -              & 14.36                 & 20.88                 \\
        DVSE~\cite{liu2015multi}                   & V                  & -              & -              & -              & -              & -              & 18.75                 & 21.79                 \\
        MCN~\cite{anne2017localizing}                          & V               & 11.41          & 2.72           & 24.94          & 8.22           & 10.67          & -                     & -                     \\
        CAL~\cite{escorcia2019temporal}                    & V            & 25.49          & 11.54          & 23.40          & 7.65           & 9.89           & -                     & -                     \\
        XML~\cite{lei2020tvr}                                & V         & 41.83          & 30.35          & 44.63          & 31.73          & 32.14          & 34.49                 & 55.25                 \\
        XML+\cite{lei2020tvr}                      & V                & 46.69          & 33.46          & 47.89          & 34.67          & 34.90          & 35.38                 & 55.06                 \\ \hline
        Moment-DETR~\cite{momentdetr}  & V& 52.89$_{\pm{2.3}}$  & 33.02$_{\pm{1.7}}$ & 54.82$_{\pm{1.7}}$ & 29.40$_{\pm{1.7}}$ & 30.73$_{\pm{1.4}}$ & 35.69$_{\pm{0.5}}$ & 55.60$_{\pm{1.6}}$ \\
        QD-DETR (\textbf{Ours})                  & V                  & \textbf{62.40}$_{\pm_{1.1}}$ & \textbf{44.98}$_{\pm_{0.8}}$ & \textbf{62.52}$_{\pm_{0.6}}$ & \textbf{39.88}$_{\pm_{0.7}}$ & \textbf{39.86}$_{\pm_{0.6}}$ & \textbf{38.94}$_{\pm_{0.4}}$        & \textbf{62.40}$_{\pm_{1.4}}$    \\ \hline
        UMT~\cite{umt}             & V+A                            & 56.23          & 41.18          & 53.38          & 37.01          & 36.12          & 38.18                 & 59.99                 \\
        QD-DETR (\textbf{Ours})  & V+A & \textbf{63.06}$_{\pm_{1.0}}$ & \textbf{45.10}$_{\pm_{0.7}}$ & \textbf{63.04}$_{\pm_{0.9}}$ & \textbf{40.10}$_{\pm_{1.0}}$ & \textbf{40.19$_{\pm_{0.6}}$} & \textbf{39.04$_{\pm_{0.3}}$}        & \textbf{62.87$_{\pm_{0.6}}$}    \\ \hlineB{2.5}
        \end{tabular}
        }
\end{table*}
\endgroup

\subsection{Experimental Settings}
\paragraph{Dataset and Evaluation Metrics.}
For the evaluation, we validate the effectiveness of query-dependent source representation on QVHighlights~\cite{momentdetr}, TVSum~\cite{song2015tvsum}, Charades-STA~\cite{gao2017tall}.
\textbf{QVHighlights} is the most recently publicized dataset for both moment retrieval and highlight detection. It is also the only dataset that has annotations for both tasks.
In detail, QVHighlights consists of over 10,000 videos annotated with human-written text queries.
It provides a fair benchmark as the evaluation for the test split can only be measured through submitting the prediction to the QVHighlights server\footnote{https://codalab.lisn.upsaclay.fr/competitions/6937}.
\textbf{Charades-STA} and \textbf{TVSum} are the dataset for moment retrieval and video summarization, respectively.
Each of them contains 9,848 videos regarding indoor activities and 50 videos of various genres, e.g., news, documentary, and vlog.
For all datasets, we follow the data splits from the existing works~\cite{momentdetr, umt}.

To measure the performances, we use the same evaluation metrics used in the baselines. 
Specifically, recall@1 with IoU thresholds 0.5 and 0.7, and mean average precision~(mAP) at different thresholds.
Similarly, we use mAP and HIT@1 for evaluating the highlight detection.
HIT@1 is computed through the hit ratio of the highest-scored clip.

\begin{table*}[]
	\centering
	{\small
	\caption{Highlight detection performance comparison on TVsum dataset. }
	\label{table_TVsum}
    \begin{tabular}{l|c|cccccccccc|c}
    \hlineB{2.5}
    \multicolumn{1}{c|}{Method} & Src & VT        & VU        & GA        & MS        & PK        & PR        & FM        & BK   & BT   & DS   & Avg. \\ \hlineB{2.5}
    sLSTM~\cite{zhang2016video}   &V                  & 41.1      & 46.2      & 46.3      & 47.7      & 44.8      & 46.1      & 45.2      & 40.6 & 47.1 & 45.5 & 45.1 \\
    SG~\cite{mahasseni2017unsupervised} &V                    & 42.3      & 47.2      & 47.5      & 48.9      & 45.6      & 47.3      & 46.4      & 41.7 & 48.3 & 46.6 & 46.2 \\
    LIM-S~\cite{xiong2019less}         &V              & 55.9      & 42.9      & 61.2      & 54.0      & 60.3      & 47.5      & 43.2      & 66.3 & 69.1 & 62.6 & 56.3 \\
    Trailer~\cite{wang2020learning}    &V                 & 61.3      & 54.6      & 65.7      & 60.8      & 59.1      & 70.1      & 58.2      & 64.7 & 65.6 & 68.1 & 62.8 \\
    SL-Module~\cite{xu2021cross}       &V            & 86.5      & 68.7      & 74.9      & \textbf{86.2}      & 79.0      & 63.2      & 58.9      & 72.6 & 78.9 & 64.0 & 73.3 \\ 
    QD-DETR (\textbf{Ours}) &V    & \textbf{88.2} & \textbf{87.4} & \textbf{85.6} & 85.0 & \textbf{85.8} & \textbf{86.9} & \textbf{76.4} & \textbf{91.3} & \textbf{89.2} & \textbf{73.7} & \textbf{85.0} \\ \hline
    MINI-Net~\cite{hong2020mini}       &V+A             & 80.6      & 68.3      & 78.2      & 81.8      & 78.1      & 65.8      & 57.8      & 75.0 & 80.2 & 65.5 & 73.2 \\
    TCG~\cite{ye2021temporal}         &V+A                & 85.0      & 71.4      & 81.9      & 78.6      & 80.2      & 75.5      & 71.6      & 77.3 & 78.6 & 68.1 & 76.8 \\
    Joint-VA~\cite{badamdorj2021joint}   &V+A                 & 83.7      & 57.3      & 78.5      & 86.1      & 80.1      & 69.2      & 70.0      & 73.0 & \textbf{97.4} & 67.5 & 76.3 \\
    UMT~\cite{umt}        &V+A          & 87.5      & 81.5      & 88.2      & 78.8      & 81.4      & 87.0      & 76.0      & 86.9 & 84.4 & \textbf{79.6} & 83.1 \\ \hline
    QD-DETR (\textbf{Ours}) &V+A & \textbf{87.6} & \textbf{91.7} & \textbf{90.2} & \textbf{88.3} & \textbf{84.1} & \textbf{88.3} & \textbf{78.7} & \textbf{91.2} & 87.8 & 77.7 & \textbf{86.6}  \\ \hlineB{2.5}
    \end{tabular}
    }
    \vspace{-0.1cm}  %
\end{table*}

\begingroup
\setlength{\tabcolsep}{2pt} 
\renewcommand{\arraystretch}{1} 
\begin{table}[]
    \centering
    {\footnotesize
    \caption{Charades dataset. $\dagger$ denotes the method using the video and audio as the source. SF+C stands for Slowfast and CLIP features.}
    \label{table_charades}
    \begin{tabular}{llll|llll}
    \hlineB{2.5}
    Method & feat & R1@0.5 & R1@0.7 & Method      & feat & R1@0.5 & R1@0.7 \\ \hlineB{2.5}
    SAP    & VGG  & 27.42  & 13.36  & CTRL        & C3D  & 23.63  & 8.89   \\
    TripNet& VGG  & 36.61  & 14.50  & ACL         & C3D  & 30.48  & 12.20  \\
    SM-RL  & VGG  & 24.36  & 11.17  & RWM-RL      & C3D  & 36.70  & -      \\
    MAN    & VGG  & 41.24  & 20.54  & MAN         & C3D  & 46.53  & 22.72  \\
    2D-TAN & VGG  & 40.94  & 22.85  & DEBUG       & C3D  & 37.39  & 17.69  \\
    FVMR   & VGG  & 42.36  & 24.14  & VSLNet       & C3D  & 47.31  & 30.19  \\
    UMT$\dagger$  & VGG  & 48.31  & 29.25  & \textbf{Ours}        & C3D  & \textbf{50.67}  & \textbf{31.02}  \\ \cline{5-8} 
    \textbf{Ours}   & VGG  & 52.77  & 31.13  & M-DETR & SF+C & 53.63  & 31.37  \\
    \textbf{Ours}$\dagger$ & VGG  & \textbf{55.51}  & \textbf{34.17}  & \textbf{Ours}        & SF+C & \textbf{57.31}  & \textbf{32.55}  \\ \hlineB{2.5}
    \end{tabular}
    }
\end{table}
\endgroup

\subsection{Experimental Results}
We compare QD-DETR against baselines in MR and HD
throughout Tab.~\ref{table_QVHighlight}, Tab.~\ref{table_charades}, and Tab.~\ref{table_TVsum}. 
Our experiments with multi-modal sources, i.e., video with audio, are implemented by simply concatenating the video and audio along the channel axis.
Throughout the tables, we use bolds to denote the best scores.

In Tab.~\ref{table_QVHighlight}, the task is to jointly learn and predict MR/HD.
As observed, our QD-DETR outperforms state-of-the-art~(SOTA) approaches with all evaluation metrics.
Among methods utilizing the video source, QD-DETR shows a dramatic increase with stricter metrics with high IOU; it outperforms previous SOTA by large margins up to 36\% in R1@0.7 and mAP@0.75.
On the other hand, QD-DETR with video and audio sources boosts 11.84\% on average of the metrics for MR compared to the SOTA method employing the multi-modal source data.
These results verify the importance of emphasizing the source~(video-only or video+audio) descriptive contexts in the text queries.

\begin{table*}[t]
	\centering
	{\small
	\vspace{0.1cm}  %
	\caption{Ablation study on QVHighlights $\textit{val}$ split. CATE and DAM stands for cross-attentive transformer encoder and using dynamic anchor moments as the decoder query, respectively. All the quantities are averaged over 5 runs.}
	\label{table_ablation}
        \begin{tabular}{c|c|c|c|c|ccccccc}
        \hlineB{2.5}
        \multicolumn{1}{c|}{\multirow{3}{*}{}} & \multicolumn{1}{c|}{\multirow{3}{*}{CATE}} & \multicolumn{1}{c|}{\multirow{3}{*}{Neg. Pair}} & \multicolumn{1}{c|}{\multirow{3}{*}{Saliency Token}} & \multicolumn{1}{c|}{\multirow{3}{*}{DAM}} & \multicolumn{5}{c}{MR}                                                             & \multicolumn{2}{c}{HD}                        \\ \cline{6-12} 
        \multicolumn{1}{c|}{} & \multicolumn{1}{c|}{} & \multicolumn{1}{c|}{} & \multicolumn{1}{c|}{} & \multicolumn{1}{c|}{} & \multicolumn{2}{c}{R1}          & \multicolumn{3}{c}{mAP}                          & \multicolumn{2}{c}{\textgreater{}= Very Good} \\ \cline{6-12} 
        
        \multicolumn{1}{c|}{} &\multicolumn{1}{c|}{} & \multicolumn{1}{c|}{} & \multicolumn{1}{c|}{} & \multicolumn{1}{c|}{} & @0.5           & @0.7           & @0.5           & @0.75          & Avg.           & mAP                   & HIT@1                 \\ \hline

        (a) &  &  &  &  & 52.89 & 33.02 & 54.82 & 29.40 & 30.73 & 35.69 & 55.60 \\ \hline
        (b) &\cmark &  &  &  & 56.16 & 38.71 & 56.48 & 33.42 & 34.07 & 37.14 & 58.34 \\ 
        (c) && \cmark &  &  & 58.69 & 39.83 & 58.39 & 34.84 & 35.40 & 39.02 & 62.81 \\
        (d) &&  & \cmark &  & 55.48 & 37.00 & 55.81 & 26.75 & 32.84 & 37.48 & 58.59 \\ 
        (e) &&  &  & \cmark & 53.19 & 35.91 & 55.58 & 32.55 & 33.33 & 35.68 & 55.56 \\ \hline
        (f) & \cmark &  &  & \cmark & 57.72 & 42.35 & 59.10 & 38.16 & 38.03 & 36.56 & 57.44 \\
        (g) & \cmark & \cmark & &  & 59.57 & 42.12 & 59.19 & 36.63 & 36.76 & 38.64 & 61.62 \\ 
        (h) && \cmark & \cmark &  & 60.00 & 40.97 & 59.21 & 35.41 & 35.89 & 39.06 & 62.88 \\ 
        (i) &\cmark & \cmark & \cmark &  & 60.32 & 42.39 & 59.47 & 36.79 & 36.93 & \textbf{39.21} & 62.76 \\ \hline
        (j) &\cmark  & \cmark & \cmark & \cmark & \textbf{62.68} & \textbf{46.66} & \textbf{62.23} & \textbf{41.82} & \textbf{41.22} & 39.13 & \textbf{63.03} \\ \hlineB{2.5}
        \end{tabular}
        }
\end{table*}

Results in Tab.~\ref{table_charades} also compare MR performances against the models using VGG~\cite{semanticprop, wang2019language, zhang2019man, zhang2020learning, gao2021fast, umt},  C3D~\cite{gao2017tall, liu2018attentive, he2019read, zhang2019man, lu2019debug, zhang2020span}, and Slowfast~(SF) and CLIP features~\cite{momentdetr} on Charades dataset.
For a fair comparison, we enumerate each method with its backbone and compare within it.
For each feature from VGG, C3D, and SF+C, we follow the data preparation settings from UMT~\cite{umt}, VSLNet~\cite{zhang2020span}, and Moment-DETR~\cite{momentdetr}.
As reported, we validate that our model surpasses the existing SOTA methods in every type of feature.

\begin{table*}[]
	\centering
	{\small
	\vspace{0.1cm}  %
 \caption{Ablation study on cross-attention transformer encoder. 
    We compare ours against the deepened transformer encoder with self-attention layers to validate that the performance gain does not come from additional parameters.
	SATE and CATE each indicate the transformer encoder only with self-attention layers and our transformer encoder. The numbers in the parenthesis denote the number of layers. For the experiment with $\dagger$, we only use the query features as the condition in the encoder and only the video representations are processed by the decoder.}
	\label{table_ablation_T2V_4layer}
        \begin{tabular}{c|ccccccc}
        \hlineB{2.5}
        \multicolumn{1}{c|}{\multirow{3}{*}{T2V}} &  \multicolumn{5}{c}{MR} & \multicolumn{2}{c}{HD}                        \\ \cline{2-8} 
        \multicolumn{1}{c|}{} & \multicolumn{2}{c}{R1}          & \multicolumn{3}{c}{mAP}                          & \multicolumn{2}{c}{\textgreater{}= Very Good} \\ \cline{2-8} 
        \multicolumn{1}{c|}{}  & @0.5           & @0.7           & @0.5           & @0.75          & Avg.           & mAP                   & HIT@1                 \\ \hline
        Moment-DETR~(SATE 2)  &  52.89$_{\pm{2.3}}$  & 33.02$_{\pm{1.7}}$ & 54.82$_{\pm{1.7}}$ & 29.40$_{\pm{1.7}}$ & 30.73$_{\pm{1.4}}$ & 35.69$_{\pm{0.5}}$ & 55.60$_{\pm{1.6}}$ \\
        Moment-DETR~(SATE 4)  &  53.60$_{\pm{1.2}}$  & 35.81$_{\pm{0.9}}$ & 54.55$_{\pm{0.8}}$ & 30.64$_{\pm{0.7}}$ & 31.74$_{\pm{0.4}}$ & 35.96$_{\pm{0.2}}$ & 56.56$_{\pm{0.9}}$ \\
        Moment-DETR~(CATE 4)  &  55.10$_{\pm{0.7}}$  & 37.02$_{\pm{0.9}}$ & 56.21$_{\pm{0.3}}$ & 32.00$_{\pm{0.9}}$ & 33.19$_{\pm{0.6}}$ & 36.43$_{\pm{0.3}}$ & 56.98$_{\pm{0.6}}$ \\ 
        Moment-DETR~(CATE 4)$\dagger$  &  56.16$_{\pm{1.2}}$ & 38.71$_{\pm{1.1}}$ & 56.48$_{\pm{0.8}}$ & 33.42$_{\pm{0.7}}$ & 34.07$_{\pm{0.6}}$ & 37.14$_{\pm{0.4}}$ & 58.34$_{\pm{0.4}}$ \\  \hline
        QD-DETR~(SATE 4)$\dagger$ & 60.48$_{\pm_{0.7}}$ & 45.21$_{\pm_{1.0}}$ & 60.84$_{\pm_{0.5}}$ & 40.45$_{\pm_{0.7}}$ & 40.12$_{\pm_{0.6}}$ & 38.66$_{\pm_{0.2}}$ & 61.29$_{\pm_{1.0}}$    \\
        QD-DETR~(CATE 4)$\dagger$ &  \textbf{62.68$_{\pm_{1.1}}$} & \textbf{46.66$_{\pm_{0.6}}$} & \textbf{62.23$_{\pm_{1.0}}$} & \textbf{41.82$_{\pm_{0.9}}$} & \textbf{41.22$_{\pm_{0.4}}$} & \textbf{39.13$_{\pm_{0.3}}$}        & \textbf{63.03$_{\pm_{0.5}}$}    \\  \hlineB{2.5}
        
        \end{tabular}
        }
\end{table*}


For video highlight detection in Tab.~\ref{table_TVsum}, we follow the protocols from the previous work~\cite{umt}. 
Specifically, we train the model for each category and average the mAP scores.
Out of 10 categories, QD-DETR outperforms baselines on 9 categories when the only video source is available, and 8 categories when both video and audio are available. 
Overall, compared to methods with video-only and multi-modal sources, QD-DETR establishes new SOTA performances by improving by 4.2\% in average compared to the previous SOTA model.

\begin{figure}
    \centering
    \includegraphics[width=1\linewidth]{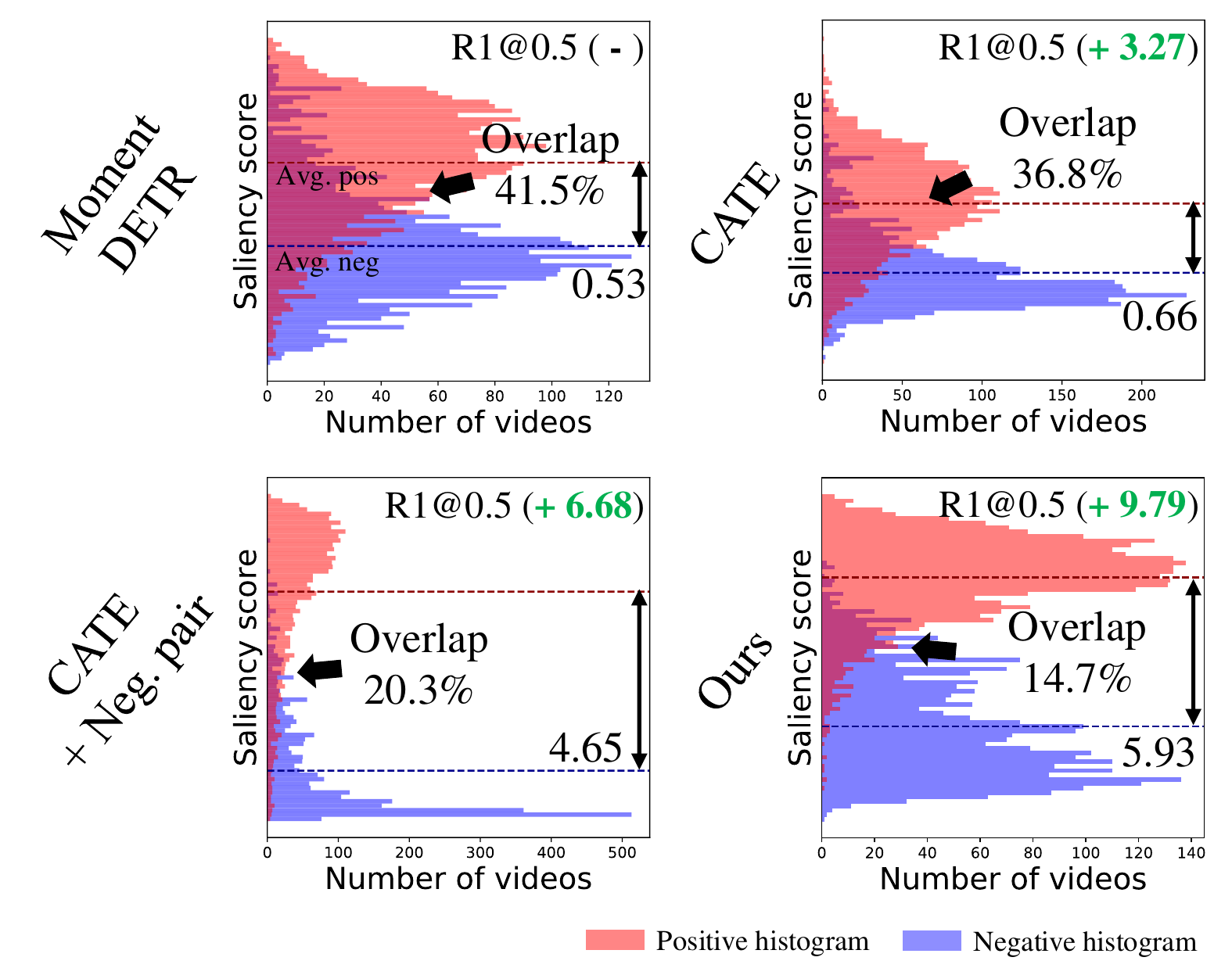}
    \vspace{-0.8cm}
    \caption{
        Ablation study in terms of saliency scores.
        We plot the histograms of the average value of saliency scores in each video when the positive and negative text queries are given. 
        For positive scores, we only account the scores within GT moments.
        The average value of each histogram are visualized by the dotted line.
        The decrease in the overlap between histograms and the increase in the gap between average values confirms the gradual improvements of significance of the query in extracting video representation.
    }
    \vspace{-0.6cm}
    \label{fig:saliency_stat}
\end{figure}

\begin{figure*}
    \centering
    \includegraphics[width=1.\linewidth]{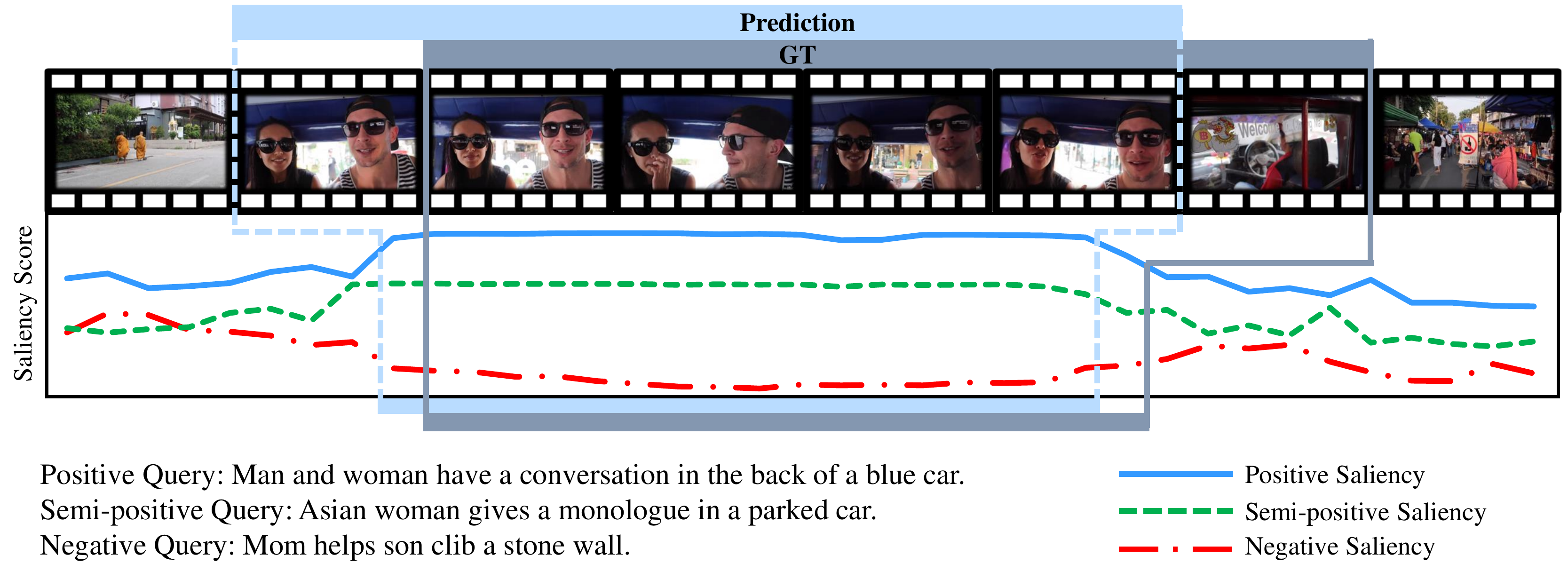}
    \vspace{-0.5cm}
    \caption{
    Visualization of results predicted by QD-DETR. 
    Predicted and ground-truth moments are bounded by the lines. 
    Blue, green, and red lines indicate the saliency scores for positive, semi-positive, and negative queries. The positive saliency scores are consistently higher than the others, while the scores for semi-positive are higher than the ones for the negative.
    }
    \label{fig:qualitative}
    \vspace{-0.3cm}
\end{figure*}



\subsection{Ablation study}

To investigate the effectiveness of each component in our work, we conduct an extensive ablation study in Tab.~\ref{table_ablation}.
Note that, CATE and DAM denote cross-attentive transformer encoder and dynamic anchor moments, respectively.
Rows (b) to (e) show the effectiveness of each component compared to the baseline~(a).
To explain, whereas (e) only boosts the MR performances since it only affects the transformer decoder, (b), (c), and (d) are especially beneficial for both MR/HD tasks since they are focused on query-dependent video representations 
; (b) ensures the contributions of text query in the video representation, (c) fully exploits the contexts of the text query, and (d) provides input-adaptive saliency predictor instead of MLP. 
Moreover, while our components are verified that they are all complementary to others, DAM's effectiveness is especially dependent on the usage of CATE~(compare between \{(a, e)\} and \{(b, f), (i, j)\}).
We claim that this is because DAM exploits the position information of the input tokens to capture the corresponding moments.
However, without CATE, input tokens are a mixture of multi-modal tokens, thereby providing confusing position information.

To provide in-depth examinations of each component, we inspect the difference between the positive and the negative saliency scores in Fig.~\ref{fig:saliency_stat}.
Since the role of text query is trivial in our baseline, each distribution significantly overlies on top of the other. 
Then, as we add CATE and negative pair learning, we observe a consistent decrease in overlapped areas and a larger gap between the average saliency scores of  positive and negative histograms.
Also, we believe that the widely-distributed histogram of saliency scores for the 'CATE+Neg.pair' is due to using an identical criterion for saliency prediction for diverse video-query representations.
By employing an input-adaptive saliency predictor, we notice that scores for positive queries are in almost optimal shape.

In addition, some might ask whether CATE benefits the training because of additional encoder layers.
To answer this, we conduct another ablation study in Tab.~\ref{table_ablation_T2V_4layer}.
Briefly, since our transformer encoder utilizes 2 cross-attention layers and 2 self-attention layers, we conduct comparisons against the transformer encoder composed of 4 self-attention layers~(SATE).
First, we compare CATE and SATE on Moment-DETR; by comparing the results in $2^{\text{nd}}$ and $3^{\text{rd}}$ rows, we find that CATE is much more beneficial than SATE even with the same number of layers.
Furthermore, the last two rows show comparisons within QD-DETR architecture that has the same tendency.
These results clearly demonstrate that the improvements from CATE are mainly from emphasizing the role of the text query rather than additional layers.


\subsection{Qualitative Results}
In this subsection, we study how the query-dependent video representation sensitively reacts to the change in the contexts of the text query.
In Fig.~\ref{fig:qualitative}, the measured saliency scores according to the video-query relevance are visualized.
We found that the more the query is relevant to the video clips, the higher the saliency scores retained for the query.
For instance, whereas the negative query that is totally irrelevant to the video instance has the lowest scores, the scores for semi-positive reside between the positive and the negative ones.  
Also, we find that QD-DETR sometimes provides a more precise moment prediction than a given ground-truth moment, as can be seen with the temporal box bounded by the dotted lines.
We believe that the tendency of a bit higher saliency scores at non-relevant clips for a positive query is due to the information mixing in the self-attention layers.

\section{Limitation and Conclusion}
\paragraph{Limitation}
As elaborated in the paper, we aim to highlight the role of the text query in retrieving the relevant moments and estimating their accordance level with the given text query. 
Likewise, the proposed components expect a given query to maintain a meaningful context. 
If not, and noisy text queries are provided, i.e., mismatched or irrelevant ground truth texts, the training may not be effective as reported.
\vspace{-10pt}
\paragraph{Conclusion}
Although the advent of transformer architecture has been powerful for MR/HD, investigation of the role of text query has been lacking in such architectures.
Therefore, we focused on studying the role of the text query.
As we found that the textual information is not fully exploited in expressing the video representations, we designed the cross-attentive transformer encoder and proposed a negative-pair training scheme.
Cross-attentive encoder assures the query's contributions while extracting video representation, and negative-pair training enforces the model to learn the relationship between query and video by preventing solving the problems without consideration of the query.
Finally, to preserve the diversity of query-dependent video representation, we defined the saliency token to be an input-adaptive saliency predictor.
Extensive experiments validated the strength of QD-DETR with superior performances. 

\vspace{5pt}
\noindent\textbf{Acknowledgements.} This work was supported in part by MSIT/IITP (No. 2022-0-00680, 2019-0-00421, 2020-0-01821, 2021-0-02068), and MSIT\&KNPA/KIPoT (Police Lab 2.0, No. 210121M06).
\section{Training Details}

In this section, we elaborate on the implementation details and hyperparameters used for experiments in the main manuscript.
To unify configurations across all experiments, our encoder composes of 4 layers of transformer block~(2 cross-attention layers and 2 self-attention layers) whereas there are only 2 layers in the decoder~(For HD dataset, i.e., TVSum, we only use encoding layers).
We set the hidden dimension of transformers as 256, and use the Adam optimizer with a weight decay of 1e-4.
Besides, we set the temperature of a scaling parameter $\tau$ for contrastive loss as 0.5 for all experiments.
Loss balancing parameters are $\lambda_{\text{margin}}=1$, $\lambda_{\text{cont}}=1$, $\lambda_{L1}=10$, $\lambda_{\text{gIoU}}=1$, $\lambda_{\text{CE}}=4$ and $\lambda_{\text{neg}}=1$, unless otherwise mentioned.
Additionally, we use the PANN~\cite{kong2020panns} model trained on AudioSet~\cite{gemmeke2017audio} to extract audio features\footnotemark[1] for experiments with the audio modality.

Other configurations are described as follows:

\noindent\textbf{QVHighlight.}
We use video features extracted from both pretrained SlowFast~\cite{slowfast}~(SF) and CLIP encoder~\cite{CLIP}, and text embeddings from CLIP, following the Moment-DETR.
We train QD-DETR for 200 epochs with a batch size of 32 and a learning rate of 1e-4.

\noindent\textbf{Charades-STA.}
We utilize official VGG~\cite{VGG} features with GloVe~\cite{pennington2014glove} text embedding.
To compare with additional baselines, we also test our model on pretrained C3D~\cite{C3D}, SlowFast and CLIP for video features with CLIP text embedding.
Specifically, we utilize pre-extracted features provided by other baselines repositories: UMT\footnote{https://github.com/TencentARC/UMT}, VSLNet\footnote{https://github.com/IsaacChanghau/VSLNet} and Moment-DETR\footnote{https://github.com/jayleicn/moment\_detr}.
We train ours for 100 epochs with a batch size of 8 and a learning rate of 1e-4.

\noindent\textbf{TVSum.}
I3D~\cite{I3D} features pretrained on Kinetics-400~\cite{kay2017kinetics} are utilized as a visual one, and CLIP features are used for the text embedding.
Following the most recent work~\cite{umt}, we train our model for 2000 epochs with a learning rate of 1e-3. The batch size is set to 4.
\section{Further study on model performance on varying lengths of the query.}

As discussed in the limitation, the performance of QD-DETR may depend on the quality of provided ground truth text descriptions.
Yet, this does not imply the QD-DETR's vulnerability against commonly used meaningless words in text descriptions.
As we think the queries with longer lengths may have a higher chance of including noisy texts, we divide the validation set into 3 groups each with long-, medium-, and short-length queries, and report the query-length-wise performances of QD-DETR in \cref{table_CATE_baselines}.
As shown, QD-DETR works well regardless of the query length, showing [36.7, 28.0, 26.3\%] and [7.3, 11.8, 11.1\%] improvements in mAP each for MR and HD with [Short, Medium, Long] queries.
This study implies that while irrelevant~(wrong) text descriptions for video contexts can degrade the effectiveness of QD-DETR, QD-DETR is robust against meaningless words that are commonly present in text queries.


\begingroup
\setlength{\tabcolsep}{4.3pt} 
\renewcommand{\arraystretch}{1} 
\begin{table}[t]
	\centering
	{\scriptsize
	\caption{Experimental results on QVHighlights.}
	\label{table_CATE_baselines}
        \vspace{-0.3cm}
        \begin{tabular}{cc|ccccccc}
        \hlineB{2.5}
        \multicolumn{2}{c|}{\multirow{3}{*}{}} &  \multicolumn{5}{c}{MR} & \multicolumn{2}{c}{HD}                        \\ \cline{3-9} 
        \multicolumn{2}{c|}{} & \multicolumn{2}{c}{R1}          & \multicolumn{3}{c}{mAP}                          & \multicolumn{2}{c}{\textgreater{}= Very Good} \\ \cline{3-9} 
        \multicolumn{2}{c|}{}  & @0.5           & @0.7           & @0.5           & @0.75          & Avg.           & mAP                   & HIT@1                 \\ \hlineB{2.5}
        \multicolumn{9}{c}{Performances with respect to query length} \\
        \multicolumn{9}{c}{S: \# words $\leq$ 8,\quad  M: 8 $<$ \# words $\leq$ 13, \quad  L: 13 $<$ \# words} \\
        \hline
        \multicolumn{1}{c|}{\multirow{2}{*}{S}}  & M-DETR & 51.82 & 34.49 & 51.48 & 29.48 & 29.43 & 37.11 & 59.27    \\ 
         \multicolumn{1}{c|}{}& QD-DETR & 63.95 & 48.18 & 61.18 & 40.93 & 40.23 & 38.67 & 63.60    \\ \hline
        \multicolumn{1}{c|}{\multirow{2}{*}{M}} & M-DETR & 57.47 & 39.22 & 57.41 & 33.43 & 34.73 & 37.49 & 56.26    \\ 
         \multicolumn{1}{c|}{}& QD-DETR & 65.91 & 51.43 & 65.48 & 45.54 & 44.46 & 40.07 & 62.90    \\ \hline
        \multicolumn{1}{c|}{\multirow{2}{*}{L}} & M-DETR & 49.35 & 32.90 & 52.89 & 29.14 & 30.54 & 35.95 & 55.16    \\ 
         \multicolumn{1}{c|}{}& QD-DETR & 57.42 & 40.32 & 61.03 & 37.67 & 38.56 & 39.24 & 61.29    \\ \hlineB{2.5} 
        \end{tabular}
        }
\end{table}
\endgroup

{\small
\bibliographystyle{ieee_fullname}
\bibliography{egbib}
}

\end{document}